\documentclass[10pt,twocolumn,letterpaper]{article}

\usepackage{cvpr}              

\usepackage{graphicx}
\usepackage{amsmath}
\usepackage{amssymb}
\usepackage{booktabs}
\usepackage{multirow}
\usepackage[misc]{ifsym}
\usepackage[accsupp]{axessibility}  
\newcommand{\ZZH}[1]{{\textcolor{black}{#1}}}
\newcommand{\HL}[1]{{\textcolor{black}{#1}}}
\newcommand{\ZCY}[1]{{\textcolor{black}{#1}}}

\usepackage[pagebackref,breaklinks,colorlinks]{hyperref}

\usepackage[capitalize]{cleveref}
\crefname{section}{Sec.}{Secs.}
\Crefname{section}{Section}{Sections}
\Crefname{table}{Table}{Tables}
\crefname{table}{Tab.}{Tabs.}

\def\cvprPaperID{5266} 
\def\confName{CVPR}
\def\confYear{2022}

\begin{document}

\title{Spatio-Temporal Gating-Adjacency GCN for Human Motion Prediction}

\author{Chongyang Zhong\textsuperscript{1,2}, Lei Hu\textsuperscript{1,2}, Zihao Zhang\textsuperscript{1,2}, Yongjing Ye$^{1,2}$, Shihong Xia\textsuperscript{1,2}\thanks{Corresponding author.} \\ 
$^1$Institute of Computing Technology, Chinese Academy of Sciences; $^2$University of Chinese Academy of Sciences\\
{\tt\small \{zhongchongyang, hulei19z, zhangzihao, yeyongjing, xsh\}@ict.ac.cn}
}
\maketitle

\begin{abstract}
Predicting future motion based on historical motion sequence is a fundamental problem in computer vision, and it has wide applications in autonomous driving and robotics. 
Some recent works have shown that Graph Convolutional Networks(GCN) are instrumental in modeling the relationship between different joints. However, \ZZH{considering the variants and diverse action types in human motion data,} the cross-dependency of the spatio-temporal relationships will be difficult to depict 
due to the decoupled modeling strategy, which may also exacerbate the problem of insufficient generalization.
Therefore, we propose the Spatio-Temporal Gating-Adjacency GCN(GAGCN) to learn the complex spatio-temporal dependencies over diverse action types.
\ZCY{Specifically, we adopt gating networks to enhance the generalization of GCN via the trainable adaptive adjacency matrix  obtained by blending the candidate spatio-temporal adjacency matrices.}
Moreover, GAGCN addresses the cross-dependency of space and time by balancing the weights of spatio-temporal modeling and fusing the decoupled spatio-temporal features. 
Extensive experiments on Human 3.6M, AMASS, and 3DPW demonstrate that GAGCN achieves state-of-the-art performance in both short-term and long-term predictions. 
\end{abstract}
\begin{figure}[!htbp]
    \centering
    \includegraphics[width = 0.5\textwidth]{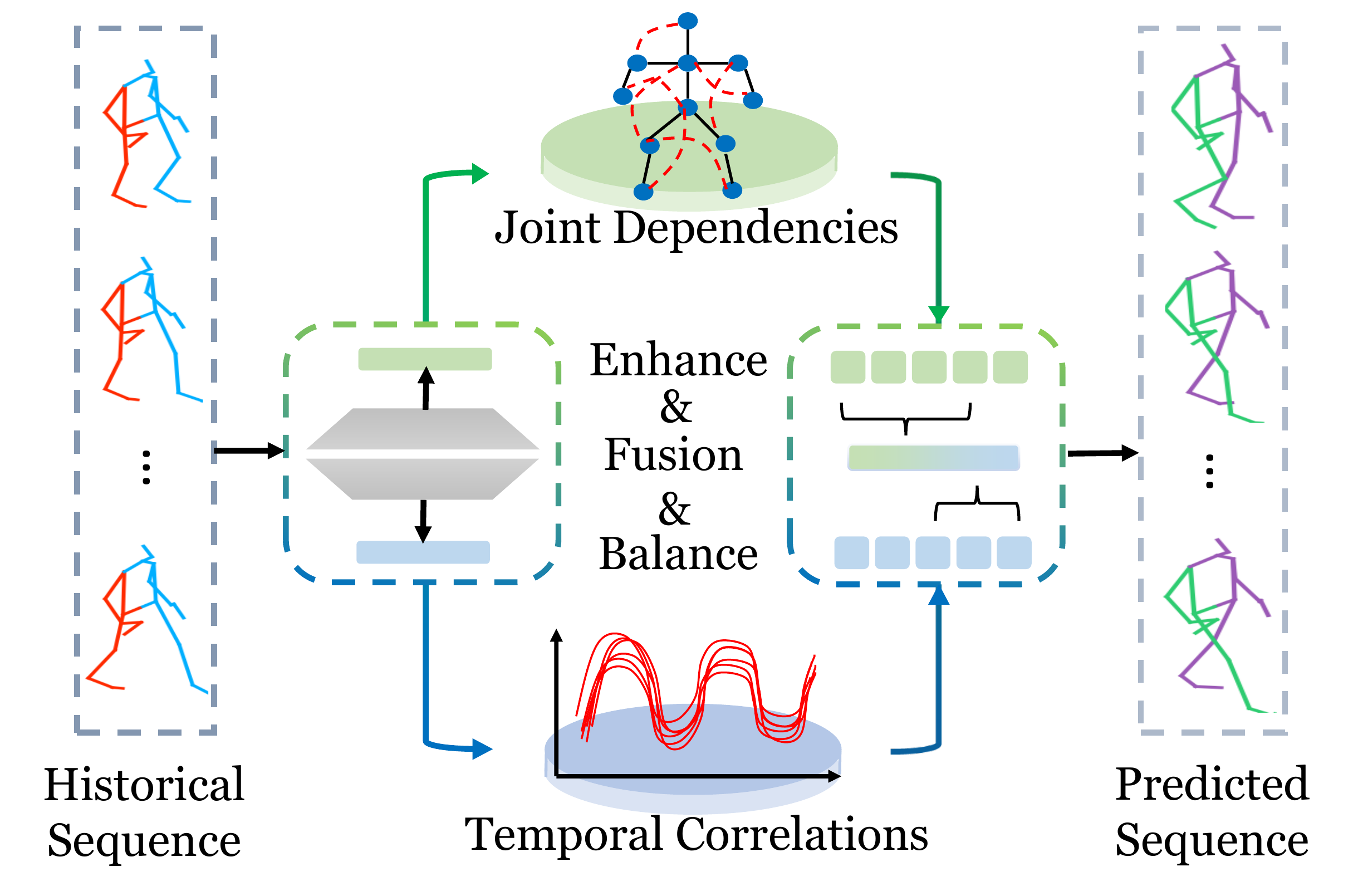}
    \caption{
    \ZZH{The illustration of our method. Given the historical input human motion sequence, we try to predict the future motion sequence by enhancing, balancing, and fusing two key factors, i.e. the joint dependencies and the temporal correlations.}}
    \label{img:01_01}
\end{figure}
\section{Introduction}
\ZZH{The aim of human motion prediction is to predict the motion trend of the skeleton-based human body in the future period from a given historical motion sequence, which is a significant computer vision task with many potential applications, such as autonomous driving, human-robotics interaction, target tracking, and motion planning. }

The skeleton-based human motion sequence is a structured time series, which means that the movement of a single joint is affected by the coupling of spatial connections with other joints and the temporal trajectory tendency. We call these complex spatio-temporal relationships as cross-dependency. \ZZH{The challenges lie in motion prediction are mainly two-fold. First, earlier literature based on Recurrent Neural Network(RNN), such as LSTM and GRU suggests that predicting the long-term sequence will meet the inherent error accumulation problem\cite{fragkiadaki2015recurrent,jain2016structural,martinez2017human,ghosh2017learning,chiu2019action,gopalakrishnan2019neural,gui2018adversarial,tang2018long,wang2019imitation}. Though subsequent convolution-based approaches\cite{hernandez2019human,butepage2017deep,li2018convolutional} for sequence-to-sequence prediction reduce the error of long-term prediction to some extent, the error accumulation is still a problem to be solved.}
Second, it is difficult to model the spatio-temporal relationships since the skeleton-based human motion is very complex and diverse. Rather than using the basic motion representation(joint angle, position, and velocity) directly or extracting the spatial features of the motion with a simple fully connected layer, recent studies try to use GCN to depict the spatio-temporal relationships\cite{mao2019learning,li2021symbiotic,mao2020history,mao2021multi,cui2020learning,li2020dynamic,li2021multiscale,sofianos2021space,liu2021motion}.

\ZCY{Though GCN-based works are instrumental for solving the long-term prediction problem to some extent, there are two issues to be explored: 1. The inter-joint and inter-frame relationships will change with the motion variance and action types, therefore a stable adjacency matrix will lead to inherent poor generalization on multi-action motions; 2. Direct concatenation of the decoupled spatial and temporal features can not fully explore the cross-dependency of the spatio-temporal relationships.}

In this paper, we propose the Spatio-Temporal Gating-Adjacency GCN(GAGCN) to learn the complex spatio-temporal dependencies over diverse action types. \ZZH{To solve the above two issues, our key idea mainly consists of two parts, namely, \textbf{the enhancing strategy} and \textbf{the balancing and fusing strategy}}(shown in Fig.~\ref{img:01_01}).
\ZCY{First, given different historical motion sequences, the gating network in our GAGCN output corresponding blending coefficients which are then used to blend the trainable candidate adjacency matrices. 
The inter-joint and inter-frame relationships of different motions are learned by the adaptive blended adjacency matrix dynamically, which enhances the generalization of our model on multi-action motions.}
Second, the proposed GAGCN can be utilized to balance the weight of spatial and temporal modeling by scaling the number of candidate matrices. And the spatio-temporal features are fused to mine the hidden cross-dependency of spatio-temporal relationships from the historical motion sequence.

Extensive experiments are conducted on Human3.6M~\cite{ionescu2013human3}, AMASS\cite{mahmood2019amass} and 3DPW\cite{von2018recovering}. We demonstrate that our method outperforms state-of-the-art methods in both short-term and long-term motion predictions. The main contributions of our work can be summarized as follows:
\begin{itemize}
    \item[1.]To the best of our knowledge, we are the first to use the gating network to enhance the generalization of GCN on human motion prediction. The adaptive adjacency matrix obtained by blending candidate matrices helps to enhance the scalability of our network across multi-action motions.
    \item[2.]We capture the cross-dependency of space and time by balancing and fusing the decoupled joint dependencies and temporal correlations to learn the more expressive embedding features.
    \item[3.]We carry out extensive experiments on Human3.6M, AMASS and 3DPW both quantitatively and qualitatively to demonstrate that the results of our method outperform state-of-the-art works.
\end{itemize}
\begin{figure*}[ht]
    \centering
    \includegraphics[width =1 \textwidth]{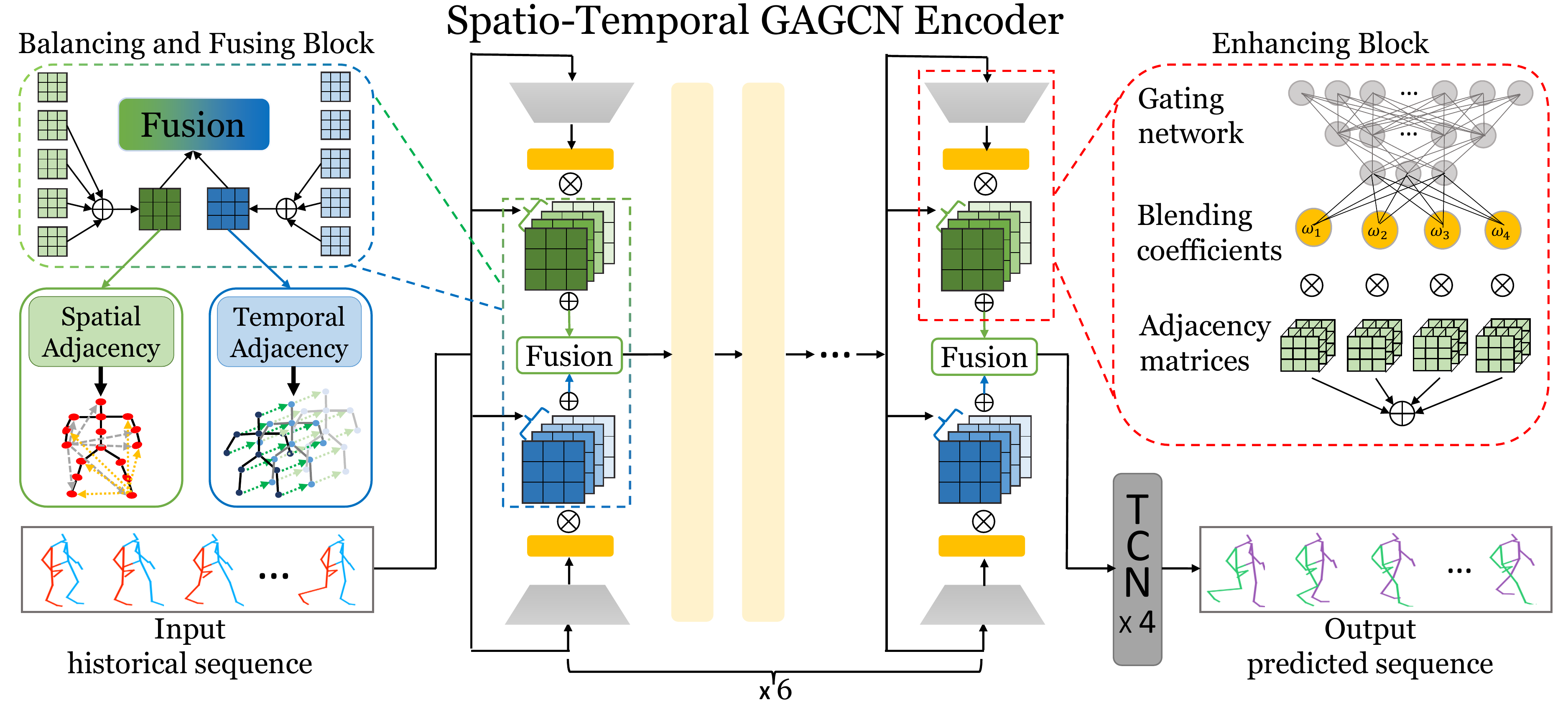}
    \caption{The overview of the proposed GAGCN network. We use Spatio-Temporal Gating-Adjacency GCN (GAGCN) as the encoder to learn the spatio-temporal dependence of the historical motion sequence, and then use TCN as a decoder. We first feed the feature from the previous layer into a spatial gating network and a temporal gating network respectively, obtaining the blending coefficients $\{w_s^i\}$ and $\{w_t^i\}$. Then we blend the spatial(temporal) adjacency matrix using the blending coefficients to create the adaptive spatial(temporal) adjacency matrix. Finally, we fuse the spatial and temporal dependencies with the Kronecker product to output the features for the next layer.}
    \vspace{-10pt}
    \label{img:01_02}
\end{figure*}
\section{Related Works} 

\textbf{Human Motion Prediction}\hspace{0.2cm}
Traditional works on motion prediction attempt to use traditional statistical methods such as hidden Markov models\cite{brand2000style} and Gaussian process hidden variable models\cite{wang2007gaussian}, which have limitations in dealing with the high-dimensional dynamics of human motion and yield unsatisfactory results. With the development of deep neural networks, exciting progress has been made in motion prediction. Some works use RNN to model the temporal correlations of human motion\cite{fragkiadaki2015recurrent,jain2016structural,martinez2017human,ghosh2017learning,chiu2019action,gopalakrishnan2019neural,gui2018adversarial,tang2018long,wang2019imitation}. However, these \HL{frame-by-frame} methods perform poorly on long-term motion prediction due to their inherent error accumulation problem, and RNN-based networks suffer from first-frame discontinuity. To address these issues, researchers have attempted to improve the prediction results of RNN-based networks using sequence-to-sequence residual models\cite{martinez2017human}, generative adversarial learning\cite{gui2018adversarial}, and imitation learning\cite{wang2019imitation}. In contrast to frame-by-frame framework, sequence-to-sequence method can effectively reduce cumulative error in long-term prediction, which includes convolution-based\cite{hernandez2019human,butepage2017deep,li2018convolutional} and attention-based mechanisms\cite{mao2020history,mao2021multi,cai2020learning}. The convolution-based approach treats the historical sequence as an entirety and extracts motion features in spatial or temporal dimensions, while the attention-based approach uses the attention model to learn the joint-to-joint and frame-to-frame dependencies. 

Recently, graph convolutional networks(GCN)\cite{DBLP:conf/iclr/KipfW17} have achieved state-of-the-art results in motion prediction\cite{mao2019learning,li2021symbiotic,mao2020history,mao2021multi,cui2020learning,li2020dynamic,li2021multiscale,sofianos2021space,liu2021motion}. Researchers use GCN with trainable adjacency matrices to model the joint dependencies of human motion. These methods learn the spatial properties of human motion by dividing the spatial properties into skeletal connections and implicit non-physical connections between individual joints\cite{cui2020learning}, providing semantic prior knowledge to the network\cite{liu2021motion}, dividing the human body into multi-scale\cite{li2020dynamic}.

Although the above works have made encouraging progress, most works treat temporal and spatial modeling in a decoupled manner and directly concatenate them while the spatial dependencies of motion are often coupled with the global temporal trajectory. To address this problem, GAGCN is proposed to learn the cross-dependency of space and time by balancing the weight of spatio-temporal modeling and fusing spatio-temporal features, which helps us to capture the spatio-temporal relationships simultaneously. 

\textbf{Spatio-Temporal Modeling for Human Motion}\hspace{0.2cm}
To the best of our knowledge, the first work to simultaneously model spatio-temporal relationships is SRNN\cite{jain2016structural}. They use a graph model to represent the human body, where the joint nodes and edge nodes are composed of RNN, thereby being the first to achieve long-term motion prediction. Another work that combines spatio-temporal modeling more closely is STGCN\cite{yan2018spatial}, where they encode both the spatial connection of human joints in a single frame and the temporal connection of the same joint between frames into a single adjacency matrix of GCN. Although their work has made impressive progress in action recognition, it is limited by the constant adjacency matrix. Recently, a newly proposed method named Space-Time-Separable GCN\cite{sofianos2021space} performs spatio-temporal modeling by factorizing the trainable adjacency matrix into temporal and spatial to achieve state-of-the-art performance in motion prediction. 

Nonetheless, considering the variants and diverse action types in human motion data, a stable adjacency matrix can not effectively capture the changing dependencies between joints and between frames, resulting in the poor generalization of GCN. 
\HL{Mixture of} Expert(MoE)\cite{jacobs1991adaptive,jordan1994hierarchical} is a traditional machine learning method that uses blending coefficients generated by a gating network to blend multiple experts. For human motion, the gating network acts as a motion classifier to automatically calculate the probability that the input motion belongs to each class of motion and blends the results of the relevant experts to obtain the optimal output, which greatly improves the generalization of multiple human motion models\cite{zhang2018mode,starke2019neural,starke2020local,starke2021neural,ling2020character}. 

Inspired by MoE, we apply the gating network on the adjacency matrices to enhance the generalization of GCN. We first adopt \HL{several candidate} adjacency matrices as experts in MoE, then we use the gating network to learn the adaptive adjacency matrix by blending the \HL{candidate} adjacency matrices according to different inputs. The adaptive adjacency matrix can capture the dynamic relationships in human motion, which is helpful to generalize across diverse action types.
\section{\HL{Our Method}}
\textbf{Problem Formulation}\hspace{0.3cm} The purpose of skeleton-based human motion prediction is to predict the future pose sequence given the historical pose sequence. We denote the historical pose sequence as $X_{1:T} = \{x_1,x_2,...,x_T\}$ with T frames, and the predicted motion sequence of future $t$ time steps as $X_{T+1:T+t} = \{x_{T+1},x_{T+2},...,x_{T+t}\}$ \ZZH{, where $x_i$ is usually represented as 3D coordinates or joint angle of the $N$ body joints.}

\textbf{Overview}\hspace{0.3cm}\ZZH{As is shown in Fig.~\ref{img:01_02},} we adopt the encoder-decoder structure to make motion prediction. \ZZH{To better retrieve the cross spatio-temporal dependency of the historical motion sequence, }we propose Gating-Adjacency GCN (GAGCN) as the encoder \ZZH{which consists of three parts}.
\ZZH{First, the features from the previous layer are fed into a spatial gating network and a temporal gating network respectively to get the blending coefficients $\{w_s^i\}$ and $\{w_t^i\}$ }.
Then we blend the spatial adjacency matrix and temporal adjacency matrix \ZZH{by using the estimated }blending coefficients to obtain the adaptive adjacency matrix. Finally, we fuse the spatial and temporal dependencies with the Kronecker product to output the features for the next layer.
\ZZH{For the decoder, given the latent motion representation after passing through 6 GAGCN layers, we use the Temporal Convolutional Networks(TCN) to predict the future sequence.}

\subsection{Review of GCN}
In recent years, GCN-based networks have been widely utilized for the modeling of spatio-temporal dependencies of the structural time series and made inspiring progress, which provides a means of human motion prediction. Specifically, we represent the skeleton-based pose as a graph $\mathcal{G}=(\mathcal{V},\mathcal{E})$, in which $\mathcal{V}$ is the joint-node set, $\mathcal{E}$ is edge set. The joint node features are 3D coordinates or joint angles and the edge is related to the adjacency matrix $A \in \mathbb{R}^{N \times N}$.

The state-of-the-art works use the trainable adjacency matrix to replace the constant adjacency matrix, which can model not only skeletal connections but also the implicit
dependencies of joints without natural connections, making GCN more powerful for learning spatial dependencies. A single layer with the trainable adjacency matrix can be expressed as:
\begin{equation}
    H^{l+1}=f(H^{l};A,W)=\sigma(AH^l W^l)
    \label{equa0302}
\end{equation}
where $A$, $H^l \in \mathbb{R}^{N \times F^l}$ and $W^l \in \mathbb{R}^{F^l \times F^{l+1}}$ are the trainable adjacency matrix, input feature and the trainable transformation matrix, respectively.
\subsection{Spatio-Temporal Gating-Adjacency GCN}
Most prior works perform decoupled modeling and direct concatenation for spatio-temporal relationships without considering their cross-dependency, which makes it difficult to accurately depict such complex spatio-temporal relationships. Additionally, since the inter-joint and inter-frame relationships will change with the motion variance and action types, a stable adjacency matrix will lead to inherent poor generalization on multi-action motions. Thus, we propose Spatio-Temporal Gating-Adjacency GCN(GAGCN) to cope with these issues.

\textbf{Gating Adjacency}\hspace{0.2cm}As shown in prior works\cite{mao2019learning,li2021symbiotic,mao2020history,mao2021multi,cui2020learning,li2020dynamic,li2021multiscale,sofianos2021space,liu2021motion}, a stable trainable adjacency matrix can handle spatio-temporal dependencies to some extent. However, the relationship will be difficult to depict when it comes to the variants and diverse action types in human motion data. Our "Enhancing Block"(shown in the right part of Fig.~\ref{img:01_02}) is motivated by the observation that the spatio-temporal relationships are changing with the diverse variances and action types. Therefore, we aim to find the adaptive spatio-temporal relationships to cope with multi-action motion prediction.

The Mixture of Experts(MoE) is a classic machine learning method, which is \HL{proven to be able to enhance the generalization of human motion models\cite{zhang2018mode,starke2019neural,starke2020local,starke2021neural,ling2020character}.} \ZZH{ }\ZZH{The gating network is regarded as a motion classifier to automatically calculates the probability of which motion class the input motion belongs to. The results of the relevant experts are blended to obtain the adaptive output. Therefore, the generalization of the human motion model is largely enhanced}. Inspired by MoE, we apply the gating network on GCN to learn the blended adjacency matrix, which is adaptive to the diverse motion variances and action types.

Different from the traditional MoE-based methods which apply weighted sum on all of each expert's network parameters, we only use the gating network to blend the adjacency matrix. This can make our network lightweight and ensure that only the feature learning process is affected while keeping the feature transferring process unaffected. Specifically, given the features from the previous layer, the gating network in our GAGCN output several blending coefficient parameters, which can be denoted as follows:
\begin{equation}
\begin{aligned}
    \{\omega^i\}&=Gating(H)=softmax(FC(H))
    \label{equa0303}
    \end{aligned}
\end{equation}
where $FC$ denotes the 3 \ZZH{fully connected layers}, $softmax$ is the activation function, $H$ is the input feature, $\{\omega^i\}$ is the set of blending coefficients.
Then the blending coefficients are used to blend the candidate trainable adjacency matrices to get the adaptive adjacency matrix:
\begin{equation}
    \mathcal{A} =\sum_i {A}^i\cdot \omega^i
     \label{equa0304}
\end{equation}
where $\{{A}^i\}$ is the set of trainable adjacency matrices and $A$ is the adaptive adjacency matrix.

\textbf{Spatio-Temporal Modeling}\hspace{0.2cm} Previous literature usually learn spatial and temporal dependencies in a decoupled manner and directly concatenate them, therefore the cross-dependency of the spatio-temporal \ZZH{information} is still not fully explored.
\ZZH{To address this issue, we propose the balancing and fusion strategy shown in the left dotted box in Fig.~\ref{img:01_02}. The key idea is to adjust the number of candidate adjacency matrices to control and balance the weight between the spatial and temporal modeling, and then fuse spatio-temporal features with the Kronecker product.}

Specifically, \ZZH{our balancing strategy is designed as follows:} At first, we divide the adjacency matrix $A$ into $A_s$ and $A_t$, as shown in the left of Fig.~\ref{img:01_02}. Following\cite{wang2021simple}, we treat all channels of a joint as a node instead of regarding each channel of a joint as a node, which can significantly reduce the size of the adjacency matrix and maintain the correlation of different channels of the same node. $A_s \in \mathbb{R}^{N \times N}$ represents the inter-dependencies between joint and joint, whether they have skeletal connections or not. Meanwhile, $A_t \in \mathbb{R}^{T \times T}$ is trained to learn the frame-to-frame dependencies in the historical sequence. 

Then, based on Equ.~\ref{equa0303} and Equ.~\ref{equa0304}, we adopt two gating networks to learn the blending coefficients for spatial \HL{and temporal respectively, then} blend the candidate adjacency matrices to obtain the adaptive adjacency matrix. We further update these two equations into the following form:
\begin{equation}
\begin{aligned}
    \{\omega^i_k\}=Gating_k(H),\quad \mathcal{A}_k=\sum_i^q {A_k^i}\cdot \omega^i_k
    \label{equa0305}
    \end{aligned}
\end{equation}
where $\mathcal{A}_k$ is the adaptive adjacency matrix , and the subscript $k\in\{s,t\}$ indicate "spatial" $s$ or "temporal" $t$. It is worth noting that the number of candidate adjacency matrices $q\in \{n, m\}$ can be adjusted, which indicates the complexity of spatial and temporal modeling. The proposed GAGCN can be utilized to balance the weight of spatial and temporal modeling like a scale by adjusting the number of candidate matrices. For example, we use $n=4$, $m=3$ on Human 3.6M and $n=6$, $m=4$ on AMASS.

\ZZH{As for the fusing strategy, }a single layer of GAGCN can be formulated as follows:
\begin{equation}
    H^{l+1}=\sigma((\mathcal{A}_s^l\otimes \mathcal{A}_t^l) H^l W^l)
    \label{equa0306}
\end{equation}
where $H^l \in \mathbb{R}^{w^l \times N \times T}$, $W^l \in \mathbb{R}^{w^l \times w^{l+1}}$, $\mathcal{A}_s^l$ and $\mathcal{A}_t^l$ are the input feature, trainable transformation matrix, adaptive spatial adjacency matrix and adaptive temporal adjacency matrix of layer $l$, respectively. $\otimes$ denote the Kronecker product. The temporal feature and spatial feature are fused by the Kronecker product to ensure that we can find the hidden cross-dependency of spatio-temporal relationships from the historical motion sequence. 

The fused features are fed into the next layer for further learning. Through 6 GAGCN layers, we extract flexible and implicit dependencies between joints and frames represented as the spatio-temporal features. Finally, the features are passed into TCN decoders to predict future sequence, which is confirmed to make better performance and less error accumulation than RNN\cite{bai2018empirical}.

\subsection{Training}
Our training process is end-to-end and supervised. With the help of the highly expressive spatio-temporal features extracted by the GAGCN encoder, our network uses a relatively simple loss function to get state-of-the-art results.

For 3D joint coordinates representation, we use MPJPE loss:
 \vspace{-5pt}
\begin{equation}
    L_{MPJPE}= \frac{1}{N \cdot t} \sum_{i=1}^t \sum_{j=1}^N {\|\widetilde{p}_{ij}-p_{ij}\|}_2
    \label{equa0307}
\vspace{-5pt}
\end{equation}
where ${p}_{ij}$ represents the predicted 3D coordinates of $j_{th}$ joint in $i_{th}$ frame, and $\widetilde{p}_{ij}$ is the corresponding ground truth.

For the angle-based representation, we use MAE loss:
 \vspace{-5pt}
\begin{equation}
    L_{MAE}= \frac{1}{N \cdot t} \sum_{i=1}^t \sum_{j=1}^N {\mid \widetilde{x}_{ij}-x_{ij}\mid}
    \label{equa0308}
 \vspace{-5pt}
\end{equation}
where ${x}_{ij}$ represents the predicted joint angle in exponential
map of $j_{th}$ joint in $i_{th}$ frame, and $\widetilde{x}_{ij}$ is the corresponding ground truth.

\begin{table*}[ht]
\footnotesize
\centering
\begin{tabular}{p{2cm}|p{0.38cm}p{0.38cm}p{0.38cm}p{0.38cm}p{0.38cm}p{0.52cm}|p{0.38cm}p{0.38cm}p{0.38cm}p{0.38cm}p{0.38cm}p{0.52cm}|p{0.38cm}p{0.38cm}p{0.38cm}p{0.38cm}p{0.38cm}p{0.52cm}}
\hline
  &\multicolumn{6}{c|}{Walking}&\multicolumn{6}{c|}{Eating}&\multicolumn{6}{c}{Smoking} \\
\hline
milliseconds&80&160&320&400&560&1000&80&160&320&400&560&1000&80&160&320&400&560&1000\\ 
\hline
Res-GRU\cite{martinez2017human} &23.2&40.9&61.0&66.1&71.6&79.1&16.8&31.5&53.5&61.7&74.9&98.0&18.9&34.7&57.5&65.4&78.1&102.1\\
ConSeq2Seq\cite{li2018convolutional} &17.7 &33.5& 56.3 &63.6 &72.2 &82.3 &11.0 &22.4 &40.7 &48.4 &61.3 &87.1 &11.6 &22.8 &41.3& 48.9& 60.0 &81.7\\ 
LTD-10-25\cite{mao2019learning} &12.6 &23.6 &39.4 &44.5 &51.8& 60.9& 7.7& 15.8& 30.5& 37.6& 50.0& 74.1& 8.4& 16.8 &32.5& 39.5& 51.3& 73.6\\ 
HRI\cite{mao2020history} &\textbf{10.0}& 19.5 &34.2 &39.8 &47.4& 58.1 &\textbf{6.4}& 14.0 &28.7 &36.2&50.0 &75.7 &\textbf{7.0} &14.9 &29.9 &36.4 &47.6 &69.5 \\ 
STSGCN *\cite{sofianos2021space}& 10.7 &16.9 &29.1 &32.9 &40.6 &51.8 &6.8 &\textbf{11.3}& 22.6 &25.4 &33.9& 52.4 &7.2 &\textbf{11.6} &22.3 &25.8 &33.6 &50.0\\
Ours *&10.3&\textbf{16.1}&\textbf{28.8}&\textbf{32.4}&\textbf{39.9}&\textbf{51.1}&\textbf{6.4}&11.5&\textbf{21.7}&\textbf{25.2}&\textbf{31.8}&\textbf{51.4}&7.1&11.8&\textbf{21.7}&\textbf{24.3}&\textbf{31.1}&\textbf{48.7}\\ 
\hline

  &\multicolumn{6}{c|}{Discussion} &\multicolumn{6}{c|}{Directions} &\multicolumn{6}{c}{Greeting} \\
\hline
milliseconds&80&160&320&400&560&1000&80&160&320&400&560&1000&80&160&320&400&560&1000\\ 
\hline
Res-GRU\cite{martinez2017human} &25.7 &47.8 &80.0 &91.3&109.5& 131.8 &21.6 &41.3& 72.1& 84.1& 101.1 &129.1 &31.2& 58.4 &96.3& 108.8& 126.1& 153.9\\
ConSeq2Seq\cite{li2018convolutional} &17.1 &34.5& 64.8 &77.6 &98.1& 129.3 &13.5& 29.0& 57.6& 69.7 &86.6 &115.8& 22.0 &45.0& 82.0 &96.0& 116.9& 147.3\\
LTD-10-25\cite{mao2019learning}  &12.2& 25.8& 53.9 &66.7& 87.6& 118.6 &9.2 &20.6 &46.9& 58.8 & 76.1 &108.8 &16.7 &33.9& 67.5& 81.6& 104.3& 140.2\\
HRI \cite{mao2020history}&10.2 &23.4& 52.1 &65.4 &86.6& 119.8 &7.4& 18.4 &44.5 &56.5 &73.9& 106.5& 13.7& 30.1& 63.8& 78.1& 101.9& 138.8\\
STSGCN *\cite{sofianos2021space}& 9.8 &\textbf{16.8} &33.4& 40.2& 53.4& 78.8& 7.4 &13.5 &\textbf{29.2}& 34.7 &47.6& 71.0 &12.4 &21.8 &42.1& 49.2& 64.8 &91.6\\
Ours *&\textbf{9.7}&17.1&\textbf{31.4}&\textbf{38.9}&\textbf{53.1}&\textbf{76.9}&\textbf{7.3}&\textbf{12.8}&30.3&\textbf{34.5}&\textbf{45.8}&\textbf{69.9}&\textbf{11.8}&\textbf{20.1}&\textbf{40.5}&\textbf{48.4}&\textbf{62.3}&\textbf{87.7}\\
\hline

  &\multicolumn{6}{c|}{Phoning} &\multicolumn{6}{c|}{Posing} &\multicolumn{6}{c}{Purchases} \\
\hline
milliseconds&80&160&320&400&560&1000&80&160&320&400&560&1000&80&160&320&400&560&1000\\ 
\hline
Res-GRU\cite{martinez2017human} &21.1 &38.9 &66.0 &76.4 &94.0 &126.4& 29.3& 56.1& 98.3& 114.3 &140.3 &183.2 &28.7& 52.4 &86.9 &100.7 &122.1& 154.0\\
ConSeq2Seq\cite{li2018convolutional}  &13.5 &26.6 &49.9 &59.9& 77.1& 114.0 &16.9 &36.7 &75.7 &92.9& 122.5 &187.4& 20.3& 41.8 &76.5 &89.9 &111.3& 151.5\\
LTD-10-25\cite{mao2019learning} &10.2 &20.2 &40.9 &50.9 &68.7& 105.1& 12.5 &27.5& 62.5 &79.6& 109.9& 171.7& 15.5& 32.3 &63.6 &77.3 &99.4& 135.9\\
HRI\cite{mao2020history}& 8.6& 18.3 &39.0& 49.2& 67.4& 105.0& 10.2 &24.2 &58.5 &75.8& 107.6 &178.2 &13.0 &29.2 &60.4& 73.9& 95.6& 134.2\\
STSGCN *\cite{sofianos2021space}& \textbf{8.2}& 13.7& 26.9 &30.9 &41.8& 66.1& \textbf{9.9}&18.0 &38.2& 45.6& 64.3& 106.4& \textbf{11.9}& 21.3& 42.0& 48.7& 63.7& 93.5\\
Ours *&8.8&\textbf{13.5}&\textbf{25.5}&\textbf{28.7}&\textbf{41.1}&\textbf{66.0}&10.1&\textbf{17.0}&\textbf{35.5}&\textbf{45.1}&\textbf{63.3}&\textbf{99.1}&\textbf{11.9}&\textbf{20.7}&\textbf{41.8}&\textbf{47.6}&\textbf{62.1}&\textbf{85.1}\\ 
\hline

  &\multicolumn{6}{c|}{Sitting} &\multicolumn{6}{c|}{Sitting Down} &\multicolumn{6}{c}{Taking Photo} \\
\hline
milliseconds&80&160&320&400&560&1000&80&160&320&400&560&1000&80&160&320&400&560&1000\\ 
\hline
Res-GRU\cite{martinez2017human} &23.8 &44.7 &78.0 &91.2 &113.7& 152.6 &31.7 &58.3 &96.7& 112.0 &138.8 &187.4& 21.9& 41.4& 74.0 &87.6 &110.6& 153.9\\ 
ConSeq2Seq\cite{li2018convolutional} &13.5& 27.0 &52.0& 63.1 &82.4 &120.7& 20.7& 40.6& 70.4 &82.7 &106.5& 150.3& 12.7 &26.0& 52.1& 63.6 &84.4& 128.1\\ 
LTD-10-25\cite{mao2019learning} &10.4& 21.4 &45.4 &57.3& 78.5 &118.8 &17.0 &33.4& 61.6& 74.4& 99.5 &144.1& 9.9& 20.5& 43.8& 55.2& 76.8& 120.2\\ 
HRI\cite{mao2020history}&9.3 &20.1 &44.3& 56.0& 76.4 &115.9& 14.9& 30.7&59.1& 72.0 &97.0& 143.6& 8.3 &18.4 &40.7& 51.5& 72.1& 115.9\\ 
STSGCN *\cite{sofianos2021space}& \textbf{9.1} &15.1 &29.9& \textbf{35.0}& 47.7& 75.2& 14.4& \textbf{23.7} &41.9& 47.9 &63.3& 94.3& \textbf{8.2}& 14.2& 29.7& \textbf{33.6} &47.0& 76.9\\
Ours *&9.3&\textbf{14.4}&\textbf{29.6}&38.5&\textbf{45.4}&\textbf{71.1}&\textbf{14.1}&24.8&\textbf{40.0}&\textbf{47.4}&\textbf{62.8}&\textbf{84.1}&8.5&\textbf{13.9}&\textbf{28.8}&35.1&\textbf{45.2}&\textbf{70.0}\\
\hline

  &\multicolumn{6}{c|}{Waiting} &\multicolumn{6}{c|}{Walking Dog} &\multicolumn{6}{c}{Average} \\
\hline
milliseconds&80&160&320&400&560&1000&80&160&320&400&560&1000&80&160&320&400&560&1000\\ 
\hline
Res-GRU\cite{martinez2017human}  &23.8 &44.2 &75.8 &87.7 &105.4& 135.4& 36.4& 64.8 &99.1 &110.6 &128.7 &164.5&25.3&46.8&78.2&89.9&108.2&139.4\\ 
ConSeq2Seq\cite{li2018convolutional} &14.6 &29.7& 58.1& 69.7& 87.3& 117.7 &27.7 &53.6 &90.7 &103.3&122.4 &162.4&16.6&33.5&62.0&73.5&92.1&126.8\\ 
LTD-10-25\cite{mao2019learning} &10.5& 21.6 &45.9 &57.1& 75.1 &106.9 &22.9& 43.5& 74.5& 86.4& 105.8& 142.2&12.6&25.5&50.6&61.9&81.1&115.8\\ 
HRI\cite{mao2020history} &8.7& 19.2& 43.4 &54.9& 74.5& 108.2& 20.1& 40.3& 73.3& 86.3 & 108.2 &146.9&10.4&22.1&46.5&57.5&76.6&112.3\\ 
STSGCN *\cite{sofianos2021space}& 8.6& 14.7& \textbf{29.6}&35.2 &47.3& 72.0& 17.6& 29.4& 52.6& 59.6& 74.7 &102.6&10.2&17.3&33.5&38.9&51.7&77.3\\
Ours *&\textbf{8.5}&\textbf{14.1}&29.8&\textbf{33.8}&\textbf{45.9}&\textbf{69.3}&\textbf{17.0}&\textbf{28.8}&\textbf{50.1}&\textbf{59.4}&\textbf{70.1}&\textbf{91.3}&\textbf{10.1}&	\textbf{16.9}&	\textbf{32.5}&	\textbf{38.5}&	\textbf{50.0}&	\textbf{72.9}\\
\hline

\end{tabular}
\caption{MPJPE error comparison for both short-term and long-term predictions on 14 action types in Human 3.6M. The best results are shown in bold. Our method outperforms all baselines on average over all time horizons. It is worth noting that our model makes larger improvements in action types that are difficult to predict, such as "Posing" and "Sitting down". Moreover, our method has significant advantages in long-term (1000ms) motion prediction. * means the error is computed over frames.}
\label{table:01}
\vspace{-5pt}
\end{table*}

\section{Experiments}

\ZZH{In this section, we evaluate the proposed motion prediction method. First, we will show the details of the used benchmark dataset and baselines in Sec.~\ref{ssec:dataset}. The quantitative comparison results with the state-of-the-art method will be given in Sec.~\ref{ssec:sota}. Then, we will analyze the main components of our method in Sec.~\ref{ssec:ablation_study}.  Finally, we will show the qualitative evaluation in Sec.~\ref{ssec:qualitative}}. The implementation details are shown in the supplementary material.

\subsection{Datasets and Baselines}
\label{ssec:dataset}
The datasets used in our experiments include Human 3.6M~\cite{ionescu2013human3}, AMASS\cite{mahmood2019amass}, and 3DPW\cite{von2018recovering}. We will introduce these 3 datasets as follows:

\textbf{Human 3.6M}\
Human 3.6M is the most used dataset in the field of motion prediction. Human 3.6M has 3.6 million 3D poses, consisting of 15 motion categories from 7 subjects. 
We down-sample the frame rate to 25Hz. Following\cite{martinez2017human,mao2020history}, we use subjects 1,6,7,8,9 for training, subject 11 for validation and subject 5 for testing.

\textbf{AMASS}\
\HL{The Archive of Motion Capture as Surface Shapes(AMASS) dataset} is a recently published human motion dataset, which gathers 18 existing mocap datasets, such as CMU, KIT, and BMLrub. 
We down-sample the frame rate to 25Hz as for Human 3.6M. Then, following\cite{mao2020history}, we select 8 datasets from AMASS for training, 4 datasets for validation and 1 dataset(BMLrub) for testing.

\textbf{3DPW}\
The 3D Pose in the Wild dataset consists of both indoor and outdoor actions, which contains 51,000 frames captured at 30Hz. We down-sample the frame rate to 25Hz as for Human 3.6M. We only use 3DPW to test the generalization of the models trained on AMASS.


\textbf{Metrics and Baselines}\
Our model can be trained on both 3D coordinates representation and angle-based representation. Thus, we evaluate the results on both 3D coordinates errors and angle errors. We adopt the MPJPE metrics for 3D coordinates representation and MAE angle error metrics for angle-based representation\footnote{It is worth noting that the evaluation metric we used is following STSGCN\cite{sofianos2021space}, that is, the average error over frames (denoted by  *), which we only found when we checked their test code on July 9, 2022.}. We compare our approach with Res-GRU\cite{martinez2017human}, ConSeq2Seq\cite{li2018convolutional}, LTD-10-25\cite{mao2019learning}, HRI\cite{mao2020history}, and STSGCN\cite{sofianos2021space} on Human 3.6M and LTD-10-25\cite{mao2019learning}, HRI\cite{mao2020history}, and STSGCN\cite{sofianos2021space} on AMASS and 3DPW. We adapt the code and the pre-trained models released by the authors to evaluate their results. Note that HRI\cite{mao2020history} takes the past 50 frames as input to predict the future 25 frames while others takes the past 10 frames as input to predict the future 25 frames.

\subsection{Comparisons with the State-of-the-art Methods}
\label{ssec:sota}
\textbf{Human 3.6M}\hspace{0.2cm}
Because of the ambiguity in the angle-based representation, most recent works use 3D coordinates to measure the accuracy of motion prediction, i.e. MPJPE. As in previous work, we predict future motion for 25 frames(1000ms) based on a 10(400ms) frames historical motion sequence. We select 14 action types from Human 3.6M and randomly select 8 sequences for each motion to calculate the average error. As in Table.~\ref{table:01}, we show the comparison of the short-term and long-term prediction of our model and the baselines on Human 3.6M.

Our method outperforms STSGCN over almost all time horizons. In particular, thanks to the adaptive adjacency matrix, our model makes larger improvements in action types that are difficult to predict, such as "Posing" and "Sitting down". Moreover, our method has significant advantages in long-term (1000ms) motion prediction. Nonetheless, there are very few time points where our method does not perform best. These time points are in short term with small prediction errors for all methods, thus it is reasonable to have a marginal error. The bottom right of Table.~\ref{table:01} is the average error of all action types, where our method performs better than all the comparative methods for whole time horizons.

Additionally, we demonstrate the average angle errors on Human 3.6M in Table.~\ref{table:02} with the same setting for MPJPE metrics. The results illustrate that our method also outperforms STSGCN in angle-based representation.

\textbf{AMASS \& 3DPW}\hspace{0.2cm}\label{sec:4.2}
We demonstrate the short-term and long-term prediction results on AMASS-BMLrub in Table.~\ref{table:03}. We train the model on 8 datasets from AMASS and use BMLrub for testing. AMASS has much more subjects and motion sequences than Human 3.6M, which is more suitable to test the generalization of the model. Our method outperforms STSGCN on AMASS, which proves that our model can indeed enhance the generalization of GCN.

The model trained on AMASS is further tested on 3DPW, and the results are shown in Table.~\ref{table:04}. The significantly better results compared with other methods provide another strong evidence of our model's generalization across different datasets.
\begin{table}[t]
\centering
\begin{tabular}{c|cccccc}
\hline
  \multicolumn{7}{c}{Human 3.6M-average} \\\hline
milliseconds&80&160&320&400&560&1000\\ \hline
Res-GRU &0.36 &0.67& 1.02 &1.15& -& -\\
conSeq2Seq &0.38 &0.68 &1.01 &1.13 &1.35 &1.82\\
LTD-10-25&0.30 &0.54 &0.86& 0.97& 1.15 &1.59 \\ 
HRI &0.27 &0.52& 0.82& 0.94 &1.14 &1.57\\ 
STSGCN *&\textbf{0.24}& 0.39& 0.59& 0.66& 0.79& 1.09\\ 
Ours *&\textbf{0.24}&\textbf{0.38}&\textbf{0.54}&\textbf{0.65}&\textbf{0.74}&\textbf{1.02}\\
\hline
\end{tabular}
\caption{Average MAE angle error comparison on Human 3.6M(note that Res-GRU\cite{martinez2017human} has no long-term prediction results). The best results are shown in bold. Our method achieves state-of-the-art prediction in angle-based representation}
\label{table:02}
\end{table}
\subsection{Ablation Study}
\label{ssec:ablation_study}
We perform ablation studies to evaluate the effect of two key component in our method, i.e. enhancing block, balancing block. The effect of fusion block can be found in the supplementary material.

\textbf{Effect of Enhancing Block}\hspace{0.2cm}
We have shown the generalization of our method across different datasets in~\ref{sec:4.2}, and then we will further show the generalization of our method across different action types. The results are shown in Table.~\ref{table:05}. We explore the generalization of our model by testing on unseen action types(Walking Together). The results in the second row of the table are significantly better than the first row, indicating that GAGCN helps to predict unseen action types. And the results in the second and third rows are very close, indicating that GAGCN performs accurate predictions on unseen action types.  
\label{sec:4.3}
\begin{table}[t]
\centering
\begin{tabular}{c|cccccc}
\hline
 \multicolumn{7}{c}{AMASS-BMLrub-average} \\\hline
milliseconds&80&160&320&400&560&1000\\ \hline
LTD-10-25 &11.0 &20.7 &37.8& 45.3 &57.2&75.2\\ 
HRI &11.3&20.7&35.7&42.0&51.7&67.2 \\
STSGCN *&\textbf{10.0} &12.5 &21.8 &24.5 &31.9 &45.5\\ 
Ours *&\textbf{10.0}&\textbf{11.9}&\textbf{20.1}&\textbf{24.0}&\textbf{30.4}&\textbf{43.1}\\ 
\hline
\end{tabular}
\caption{Average MPJPE error comparison on AMASS-BMLrub. The best results are shown in bold. Our method outperforms all the baselines, which proves that our model can indeed enhance the generalization of GCN across datasets.}
\label{table:03}
\vspace{-10pt}
\end{table}

\begin{table}[t]
\centering
\begin{tabular}{c|p{0.45cm}p{0.45cm}p{0.45cm}p{0.45cm}p{0.5cm}p{0.65cm}}
\hline
  \multicolumn{7}{c}{3DPW-average} \\\hline
milliseconds&80&160&320&400&560&1000\\ \hline
LTD-10-25&12.6 &23.2& 39.7 &46.6& 57.9&75.5\\ 
HRI &12.6&23.1&39.0&45.4&56.0&73.7 \\ 
STSGCN *&8.6 &12.8 &21.0 &24.5 &30.4 &42.3\\ 
Ours *&\textbf{8.4}&\textbf{11.9}&\textbf{18.7}&\textbf{23.6}&\textbf{29.1}&\textbf{39.9}\\ 
\hline
\end{tabular}
\caption{Average MPJPE error comparison on 3DPW. The best results are shown in bold. The significantly better results than other methods provide another strong evidence of our model's generalization.}
\vspace{-10pt}
\label{table:04}
\end{table}

\textbf{Effect of Balancing Block}\hspace{0.2cm}
To demonstrate the effect of balancing block, we set up three contrast experiments against our method(shown in Table.~\ref{table:06}).: 1. Contrast experiment 1 illustrates the necessity of using gating network both spatially and temporally. The better results of our method indicate that applying gating network in time and space simultaneously helps to model spatio-temporal dependencies more effectively. 2. Contrast experiment 2 shows that more candidate matrices are not always better. We empirically classify the motions in human 3.6M into roughly four categories of similar motions, thus four spatial candidate matrices are used. Too many candidate matrices will increase the complexity of the network and cause under-fitting. 3. Contrast experiment 3 says that the weight of spatio-temporal modeling affects the accuracy of the prediction results indeed, and that is why we balance them by adjusting the number of candidate matrices.

\begin{table}[t]
\centering
\begin{tabular}{p{1cm}p{1.3cm}|p{0.4cm}p{0.4cm}p{0.4cm}p{0.4cm}p{0.4cm}p{0.5cm}}
\hline
  \multicolumn{8}{c}{Human 3.6M-Walking Together} \\\hline
Model & $\quad$Motion&80&160&320&400&560&1000\\ \hline
S\&TGCN&$\quad$unseen &10.8&20.7&38.1&42.7&53.1&69.8 \\ 
GAGCN&$\quad$unseen&8.9&14.0&26.8&31.1&38.0&51.6\\ 
GAGCN&$\quad$seen&8.8&13.8&26.2&29.9&37.8&50.4\\ 
\hline
\end{tabular}
\caption{Ablation study for the effect of enhancing block. "GAGCN" denotes our proposed model and "S\&TGCN" denotes GCN with stable spatial and temporal adjacency matrix, and other experimental settings are the same for both models. "seen" and "unseen" denote that whether the action type(Walking Together) is seen during training or not. The results show that our method can enhance the generalization across unseen action types.}
\label{table:05}
\end{table}

\begin{table}[t]
\centering
\begin{tabular}{p{1.7cm}|p{0.9cm}|p{0.4cm}p{0.4cm}p{0.4cm}p{0.4cm}p{0.4cm}p{0.5cm}}
\hline
  \multicolumn{8}{c}{Human 3.6M-average} \\\hline
\multicolumn{2}{c|}{milliseconds}&80&160&320&400&560&1000\\ \hline
Our method&$S_4$, $T_3$&\textbf{10.1}&	\textbf{16.9}&	\textbf{32.5}&	\textbf{38.5}&	\textbf{50.0}&	\textbf{72.9}\\ \hline
\multirow{2}{*}{$\quad$CE1}&$S_4$, $T_1$ &12.5&19.9&38.4&51.3&68.6&93.9 \\ 
&$S_1$, $T_3$ &13.1&22.3&40.9&54.1&67.1&91.1\\ \hline
$\quad$CE2&$S_8$, $T_6$&11.4&18.1&33.6&42.5&53.7&76.9\\ \hline
$\quad$CE3&$S_3$, $T_4$&10.3&\textbf{16.9}&33.1&39.2& 52.1&75.3\\ 
\hline
\end{tabular}
\caption{Ablation for the effect of balancing block. $S$ and $T$ denote spatial and temporal adjacency matrix, and the subscripts indicate the number of matrices. "CE" denotes the contrast experiment. The best results are shown in bold.}
\label{table:06}
\end{table}

\begin{figure*}[t]
    \centering
    \includegraphics[height=6cm,width = 1\textwidth]{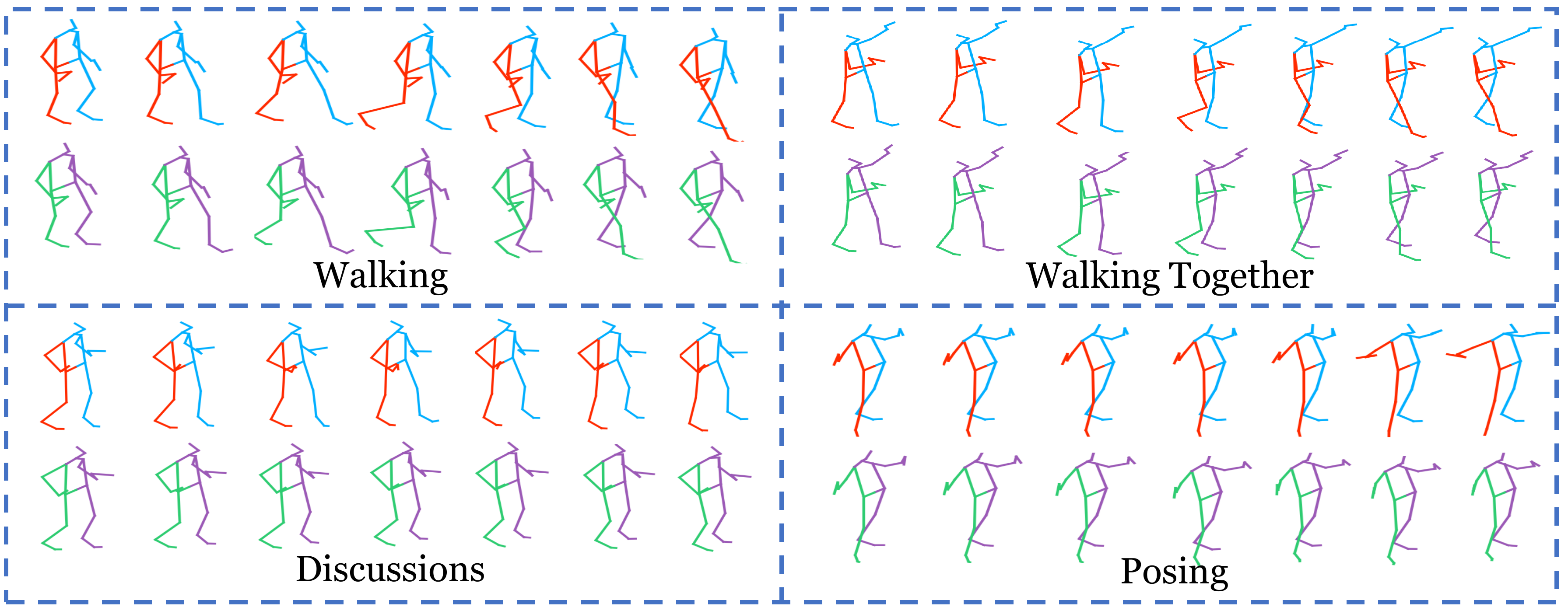}
    \caption{Visualization of predicted sequences against Ground Truth sequences for 80, 160, 320, 560, 720, 880, 1000ms. We demonstrate the prediction of "Walking", "Walking Together", "Discussions", and "Posing", where the green and purple lines indicate prediction and the red and blue lines indicate the corresponding Ground Truth. }
    \label{img:01_03}
    \vspace{-15pt}
\end{figure*}

\begin{figure}[!htbp]
    \centering
    \includegraphics[height=5cm,width = 0.5\textwidth]{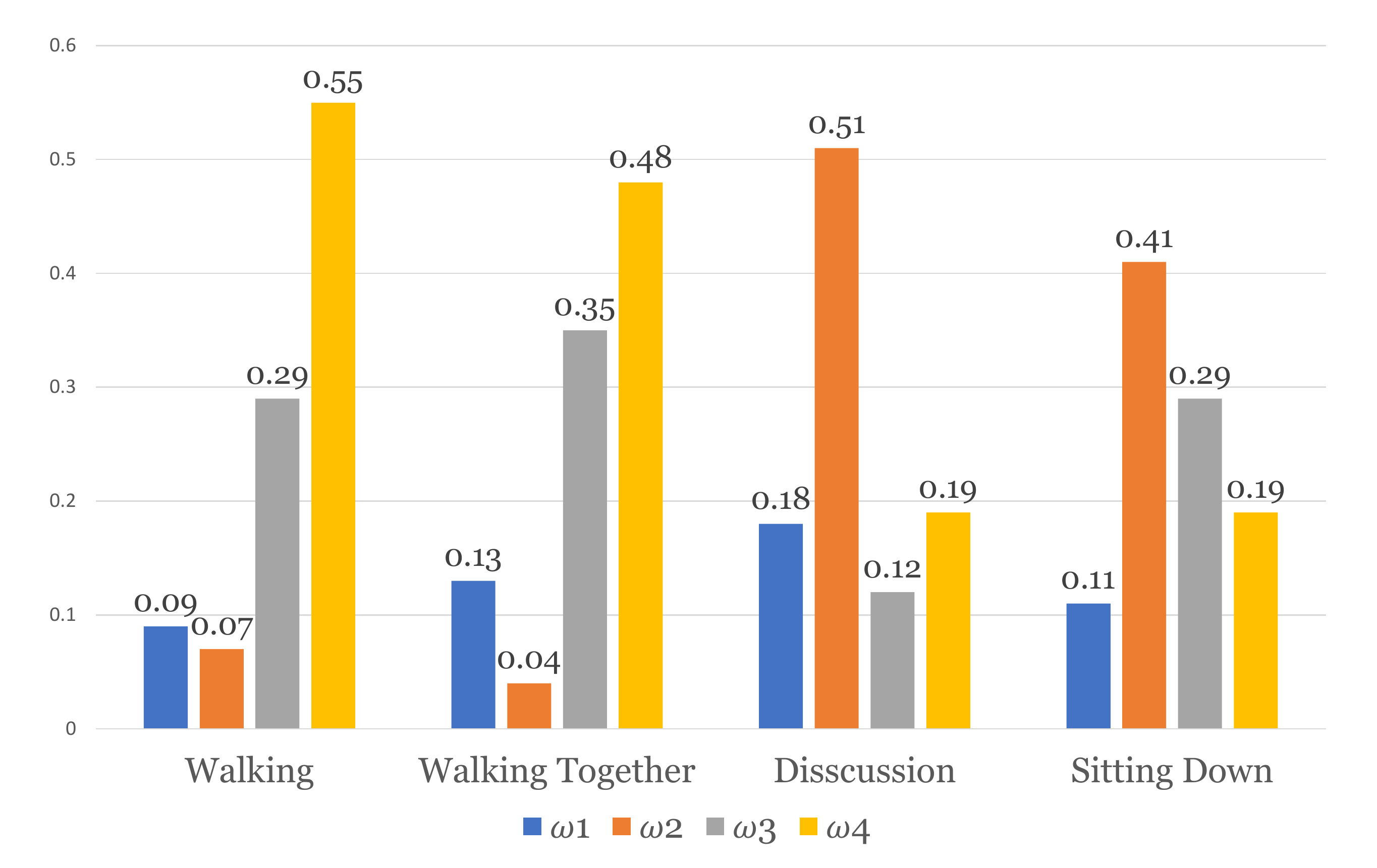}
    \caption{Visualization of average spatial blending coefficients for 4 action types. $\omega_1,\omega_2,\omega_3,\omega_4$ denote the 4 blending coefficients, respectively. Different action types(like "Walking", "Discussions" and "Sitting Down") have different coefficients distribution while the coefficients of similar motions are similarly distributed(like "Walking" and  "Walking Together").}
    \label{img:01_04}
    \vspace{-15pt}
\end{figure}

\subsection{Qualitative Evaluation}
\label{ssec:qualitative}
\textbf{Visualization of Predicted Sequence}\hspace{0.2cm}We visualize the predicted sequence on Human 3.6M and compare them with Ground Truth in Fig.~\ref{img:01_03}. For periodic motions such as "Walking" and "Walking Together", our predictions are almost identical to Ground Truth over the entire time horizons. Meanwhile, for more complex non-periodic motions like "Discussions" and "Posing", our predictions match the Ground Truth well in the short term, and the long term predictions are also quite close to GT despite some acceptable errors on the left leg of "Discussions" and the right arm of "Posing". Non-periodic motion prediction is a more challenging problem, especially when the subjects in the testing dataset perform in different ways compared with subjects in the training dataset. The visualization of predicted sequences on AMASS is shown in the supplementary material.

\textbf{Visualization of Spatial Blending Coefficients}\hspace{0.2cm}Moreover, we randomly select 16 sequences from a single action type to compute the average spatial blending coefficients(visualization of temporal blending coefficients can be found in supplementary material). Then we do the same operation on several action types and visualize them(seeing Fig.~\ref{img:01_04}). We can see that there is a clear difference in the blending coefficients distribution for different action types. The blending coefficients for "Walking Together" are derived from the partial train model in~\ref{sec:4.3}. Since "Walking Together" and "Walking" are similar periodic motions, their blending coefficients are similarly distributed with higher $\omega3$ and $\omega4$ values. That is why our model can achieve accurate prediction results for "Walking Together" without seeing it before. Non-periodic motions like "Discussion" and "Sitting Down" have higher $\omega2$ values,  but their coefficient distributions are very different. Given different inputs, GAGCN can generate the corresponding blending coefficients, which helps to learn the adaptive adjacency matrix for diverse action types. The visualization of adaptive adjacency matrix can be found in the supplementary material.
\section{Conclusion and Future Work}
In this paper, we propose a novel method called GAGCN to solve motion prediction for multi-action motions. We use the gating network to learn adaptive adjacency matrix by blending candidate adjacency matrices, which effectively enhance the generalization on multi-action motions. Meanwhile, GAGCN can balance the spatio-temporal modeling by adjusting the number of candidate matrices. Combined with the fusion of spatio-temporal features, we can extract the cross-dependency of spatial and temporal relationships to achieve state-of-the-art results on several widely used benchmark datasets. In the future, we will study how to automatically balance the weight for spatio-temporal modeling instead of manually adjusting them and explore a more efficient approach to enhance the generalization of GCN.
\\[10pt]
\hspace{-0.5cm}\textbf{\large{Acknowledgements}} 
This work was supported by the National Key R\&D Program of Science and Technology for Winter Olympics (No.2020YFF0304701) and the National Natural Science Foundation of China (No.61772499).
\newpage
{\small
\bibliographystyle{ieee_fullname}
\bibliography{ReviewTemplate}

\begin{thebibliography}{10}\itemsep=-1pt

\bibitem{bai2018empirical}
Shaojie Bai, J~Zico Kolter, and Vladlen Koltun.
\newblock An empirical evaluation of generic convolutional and recurrent
  networks for sequence modeling.
\newblock {\em arXiv preprint arXiv:1803.01271}, 2018.

\bibitem{brand2000style}
Matthew Brand and Aaron Hertzmann.
\newblock Style machines.
\newblock In {\em Proceedings of the 27th annual conference on Computer
  graphics and interactive techniques}, pages 183--192, 2000.

\bibitem{butepage2017deep}
Judith Butepage, Michael~J Black, Danica Kragic, and Hedvig Kjellstrom.
\newblock Deep representation learning for human motion prediction and
  classification.
\newblock In {\em Proceedings of the IEEE conference on computer vision and
  pattern recognition}, pages 6158--6166, 2017.

\bibitem{cai2020learning}
Yujun Cai, Lin Huang, Yiwei Wang, Tat-Jen Cham, Jianfei Cai, Junsong Yuan, Jun
  Liu, Xu Yang, Yiheng Zhu, Xiaohui Shen, et~al.
\newblock Learning progressive joint propagation for human motion prediction.
\newblock In {\em European Conference on Computer Vision}, pages 226--242.
  Springer, 2020.

\bibitem{chiu2019action}
Hsu-kuang Chiu, Ehsan Adeli, Borui Wang, De-An Huang, and Juan~Carlos Niebles.
\newblock Action-agnostic human pose forecasting.
\newblock In {\em 2019 IEEE Winter Conference on Applications of Computer
  Vision (WACV)}, pages 1423--1432. IEEE, 2019.

\bibitem{cui2020learning}
Qiongjie Cui, Huaijiang Sun, and Fei Yang.
\newblock Learning dynamic relationships for 3d human motion prediction.
\newblock In {\em Proceedings of the IEEE/CVF Conference on Computer Vision and
  Pattern Recognition}, pages 6519--6527, 2020.

\bibitem{fragkiadaki2015recurrent}
Katerina Fragkiadaki, Sergey Levine, Panna Felsen, and Jitendra Malik.
\newblock Recurrent network models for human dynamics.
\newblock In {\em Proceedings of the IEEE International Conference on Computer
  Vision}, pages 4346--4354, 2015.

\bibitem{ghosh2017learning}
Partha Ghosh, Jie Song, Emre Aksan, and Otmar Hilliges.
\newblock Learning human motion models for long-term predictions.
\newblock In {\em 2017 International Conference on 3D Vision (3DV)}, pages
  458--466. IEEE, 2017.

\bibitem{gopalakrishnan2019neural}
Anand Gopalakrishnan, Ankur Mali, Dan Kifer, Lee Giles, and Alexander~G
  Ororbia.
\newblock A neural temporal model for human motion prediction.
\newblock In {\em Proceedings of the IEEE/CVF Conference on Computer Vision and
  Pattern Recognition}, pages 12116--12125, 2019.

\bibitem{gui2018adversarial}
Liang-Yan Gui, Yu-Xiong Wang, Xiaodan Liang, and Jos{\'e}~MF Moura.
\newblock Adversarial geometry-aware human motion prediction.
\newblock In {\em Proceedings of the European Conference on Computer Vision
  (ECCV)}, pages 786--803, 2018.

\bibitem{hernandez2019human}
Alejandro Hernandez, Jurgen Gall, and Francesc Moreno-Noguer.
\newblock Human motion prediction via spatio-temporal inpainting.
\newblock In {\em Proceedings of the IEEE/CVF International Conference on
  Computer Vision}, pages 7134--7143, 2019.

\bibitem{ionescu2013human3}
Catalin Ionescu, Dragos Papava, Vlad Olaru, and Cristian Sminchisescu.
\newblock Human3. 6m: Large scale datasets and predictive methods for 3d human
  sensing in natural environments.
\newblock {\em IEEE transactions on pattern analysis and machine intelligence},
  36(7):1325--1339, 2013.

\bibitem{jacobs1991adaptive}
Robert~A Jacobs, Michael~I Jordan, Steven~J Nowlan, and Geoffrey~E Hinton.
\newblock Adaptive mixtures of local experts.
\newblock {\em Neural computation}, 3(1):79--87, 1991.

\bibitem{jain2016structural}
Ashesh Jain, Amir~R Zamir, Silvio Savarese, and Ashutosh Saxena.
\newblock Structural-rnn: Deep learning on spatio-temporal graphs.
\newblock In {\em Proceedings of the ieee conference on computer vision and
  pattern recognition}, pages 5308--5317, 2016.

\bibitem{jordan1994hierarchical}
Michael~I Jordan and Robert~A Jacobs.
\newblock Hierarchical mixtures of experts and the em algorithm.
\newblock {\em Neural computation}, 6(2):181--214, 1994.

\bibitem{DBLP:conf/iclr/KipfW17}
Thomas~N. Kipf and Max Welling.
\newblock Semi-supervised classification with graph convolutional networks.
\newblock In {\em 5th International Conference on Learning Representations,
  {ICLR} 2017, Toulon, France, April 24-26, 2017, Conference Track
  Proceedings}. OpenReview.net, 2017.

\bibitem{li2018convolutional}
Chen Li, Zhen Zhang, Wee~Sun Lee, and Gim~Hee Lee.
\newblock Convolutional sequence to sequence model for human dynamics.
\newblock In {\em Proceedings of the IEEE Conference on Computer Vision and
  Pattern Recognition}, pages 5226--5234, 2018.

\bibitem{li2021symbiotic}
Maosen Li, Siheng Chen, Xu Chen, Ya Zhang, Yanfeng Wang, and Qi Tian.
\newblock Symbiotic graph neural networks for 3d skeleton-based human action
  recognition and motion prediction.
\newblock {\em IEEE Transactions on Pattern Analysis and Machine Intelligence},
  2021.

\bibitem{li2020dynamic}
Maosen Li, Siheng Chen, Yangheng Zhao, Ya Zhang, Yanfeng Wang, and Qi Tian.
\newblock Dynamic multiscale graph neural networks for 3d skeleton based human
  motion prediction.
\newblock In {\em Proceedings of the IEEE/CVF Conference on Computer Vision and
  Pattern Recognition}, pages 214--223, 2020.

\bibitem{li2021multiscale}
Maosen Li, Siheng Chen, Yangheng Zhao, Ya Zhang, Yanfeng Wang, and Qi Tian.
\newblock Multiscale spatio-temporal graph neural networks for 3d
  skeleton-based motion prediction.
\newblock {\em IEEE Transactions on Image Processing}, 30:7760--7775, 2021.

\bibitem{ling2020character}
Hung~Yu Ling, Fabio Zinno, George Cheng, and Michiel Van De~Panne.
\newblock Character controllers using motion vaes.
\newblock {\em ACM Transactions on Graphics (TOG)}, 39(4):40--1, 2020.

\bibitem{liu2021motion}
Zhenguang Liu, Pengxiang Su, Shuang Wu, Xuanjing Shen, Haipeng Chen, Yanbin
  Hao, and Meng Wang.
\newblock Motion prediction using trajectory cues.
\newblock In {\em Proceedings of the IEEE/CVF International Conference on
  Computer Vision}, pages 13299--13308, 2021.

\bibitem{mahmood2019amass}
Naureen Mahmood, Nima Ghorbani, Nikolaus~F Troje, Gerard Pons-Moll, and
  Michael~J Black.
\newblock Amass: Archive of motion capture as surface shapes.
\newblock In {\em Proceedings of the IEEE/CVF International Conference on
  Computer Vision}, pages 5442--5451, 2019.

\bibitem{mao2020history}
Wei Mao, Miaomiao Liu, and Mathieu Salzmann.
\newblock History repeats itself: Human motion prediction via motion attention.
\newblock In {\em European Conference on Computer Vision}, pages 474--489.
  Springer, 2020.

\bibitem{mao2019learning}
Wei Mao, Miaomiao Liu, Mathieu Salzmann, and Hongdong Li.
\newblock Learning trajectory dependencies for human motion prediction.
\newblock In {\em Proceedings of the IEEE/CVF International Conference on
  Computer Vision}, pages 9489--9497, 2019.

\bibitem{mao2021multi}
Wei Mao, Miaomiao Liu, Mathieu Salzmann, and Hongdong Li.
\newblock Multi-level motion attention for human motion prediction.
\newblock {\em International Journal of Computer Vision}, pages 1--23, 2021.

\bibitem{martinez2017human}
Julieta Martinez, Michael~J Black, and Javier Romero.
\newblock On human motion prediction using recurrent neural networks.
\newblock In {\em Proceedings of the IEEE Conference on Computer Vision and
  Pattern Recognition}, pages 2891--2900, 2017.

\bibitem{sofianos2021space}
Theodoros Sofianos, Alessio Sampieri, Luca Franco, and Fabio Galasso.
\newblock Space-time-separable graph convolutional network for pose
  forecasting.
\newblock In {\em Proceedings of the IEEE/CVF International Conference on
  Computer Vision}, pages 11209--11218, 2021.

\bibitem{starke2019neural}
Sebastian Starke, He Zhang, Taku Komura, and Jun Saito.
\newblock Neural state machine for character-scene interactions.
\newblock {\em ACM Trans. Graph.}, 38(6):209--1, 2019.

\bibitem{starke2020local}
Sebastian Starke, Yiwei Zhao, Taku Komura, and Kazi Zaman.
\newblock Local motion phases for learning multi-contact character movements.
\newblock {\em ACM Transactions on Graphics (TOG)}, 39(4):54--1, 2020.

\bibitem{starke2021neural}
Sebastian Starke, Yiwei Zhao, Fabio Zinno, and Taku Komura.
\newblock Neural animation layering for synthesizing martial arts movements.
\newblock {\em ACM Transactions on Graphics (TOG)}, 40(4):1--16, 2021.

\bibitem{tang2018long}
Yongyi Tang, Lin Ma, Wei Liu, and Wei-Shi Zheng.
\newblock Long-term human motion prediction by modeling motion context and
  enhancing motion dynamics.
\newblock In {\em IJCAI}, 2018.

\bibitem{von2018recovering}
Timo von Marcard, Roberto Henschel, Michael~J Black, Bodo Rosenhahn, and Gerard
  Pons-Moll.
\newblock Recovering accurate 3d human pose in the wild using imus and a moving
  camera.
\newblock In {\em Proceedings of the European Conference on Computer Vision
  (ECCV)}, pages 601--617, 2018.

\bibitem{wang2019imitation}
Borui Wang, Ehsan Adeli, Hsu-kuang Chiu, De-An Huang, and Juan~Carlos Niebles.
\newblock Imitation learning for human pose prediction.
\newblock In {\em Proceedings of the IEEE/CVF International Conference on
  Computer Vision}, pages 7124--7133, 2019.

\bibitem{wang2021simple}
Chenxi Wang, Yunfeng Wang, Zixuan Huang, and Zhiwen Chen.
\newblock Simple baseline for single human motion forecasting.
\newblock In {\em Proceedings of the IEEE/CVF International Conference on
  Computer Vision}, pages 2260--2265, 2021.

\bibitem{wang2007gaussian}
Jack~M Wang, David~J Fleet, and Aaron Hertzmann.
\newblock Gaussian process dynamical models for human motion.
\newblock {\em IEEE transactions on pattern analysis and machine intelligence},
  30(2):283--298, 2007.

\bibitem{yan2018spatial}
Sijie Yan, Yuanjun Xiong, and Dahua Lin.
\newblock Spatial temporal graph convolutional networks for skeleton-based
  action recognition.
\newblock In {\em Thirty-second AAAI conference on artificial intelligence},
  2018.

\bibitem{zhang2018mode}
He Zhang, Sebastian Starke, Taku Komura, and Jun Saito.
\newblock Mode-adaptive neural networks for quadruped motion control.
\newblock {\em ACM Transactions on Graphics (TOG)}, 37(4):1--11, 2018.

\end{thebibliography}
}

\end{document}